%% file: adtext2image.tex
\documentclass[sigconf]{acmart}

\AtBeginDocument{%
  \providecommand\BibTeX{{%
    \normalfont B\kern-0.5em{\scshape i\kern-0.25em b}\kern-0.8em\TeX}}}

\setcopyright{acmcopyright}
\copyrightyear{2021}
\acmYear{2021}
\acmDOI{10.1145/1122445.1122456}

\acmConference[KDD '21]{KDD 2021}{August 14--18, 2021}{Virtual}
\acmBooktitle{}
\acmPrice{}
\acmISBN{}
\usepackage{amsmath}
\usepackage{algorithmicx}
\usepackage{algpseudocode}
\usepackage[]{algorithm2e}
\DeclareMathOperator*{\argmax}{arg\,max}

\setcopyright{rightsretained} 
\begin{document}

\title{VisualTextRank: Unsupervised Graph-based Content Extraction for Automating Ad Text to Image Search}


\author{Shaunak Mishra}
\affiliation{%
  \institution{Yahoo Research}
  \country{USA}
}
\email{shaunakm@verizonmedia.com}

\author{Mikhail Kuznetsov}
\affiliation{%
  \institution{Yahoo Research}
  \country{USA}
}
\email{kuznetsov@verizonmedia.com}

\author{Gaurav Srivastava}
\affiliation{%
  \institution{Yahoo Research}
  \country{USA}
}
\email{gaurav.srivastava@verizonmedia.com }

\author{Maxim Sviridenko}
\affiliation{%
  \institution{Yahoo Research}
  \country{USA}
}
\email{sviri@verizonmedia.com}


\begin{abstract}
Numerous online stock image libraries offer high quality yet copyright free images for use in marketing campaigns.
To assist advertisers in navigating such third party libraries, we study the problem of automatically fetching relevant ad images given the ad text (via a short textual query for images).
Motivated by our observations in logged data on ad image search queries (given ad text), we formulate a keyword extraction problem, where a keyword extracted from the ad text (or its augmented version) serves as the ad image query.
In this context, we propose VisualTextRank: an unsupervised method to (i) augment input ad text using semantically similar ads, and (ii) extract the image query from the augmented ad text.
VisualTextRank builds on prior work on
graph based context extraction (biased TextRank in particular) by
leveraging both the text and image of similar ads for better keyword extraction, and using advertiser category specific biasing with sentence-BERT embeddings.
Using data collected from the Verizon Media Native (Yahoo Gemini) ad platform's stock image search feature for onboarding advertisers, we demonstrate the superiority of VisualTextRank compared to competitive keyword extraction baselines (including an $11\%$ accuracy lift over biased TextRank). For the case when the stock image library is restricted to English queries, we show the effectiveness of VisualTextRank on multilingual ads (translated to English) while leveraging semantically similar English ads.
Online tests with a simplified version of VisualTextRank led to a 28.7\% increase in the usage of stock image search, and a 41.6\% increase in the advertiser onboarding rate in the Verizon Media Native ad platform.
\end{abstract}



\copyrightyear{2021}
\acmYear{2021}
\acmConference[KDD '21]{Proceedings of the 27th ACM SIGKDD Conference on Knowledge Discovery and Data Mining}{August 14--18, 2021}{Virtual Event, Singapore}
\acmBooktitle{Proceedings of the 27th ACM SIGKDD Conference on Knowledge Discovery and Data Mining (KDD '21), August 14--18, 2021, Virtual Event, Singapore}\acmDOI{10.1145/3447548.3467126}
\acmISBN{978-1-4503-8332-5/21/08}

\maketitle

\input{introduction.tex}

\input{related.tex}

\input{data.tex}
\input{method.tex}
\input{multilingual.tex}
\input{results.tex}

\input{discussion.tex}
\bibliographystyle{ACM-Reference-Format}
\bibliography{refs}
\newpage
\appendix
\input{reproducibility.tex}

\end{document}

%% file: introduction.tex
\section{Introduction} \label{sec:introduction}
\begin{figure}[]
\centering
  \includegraphics[width=1 \columnwidth]{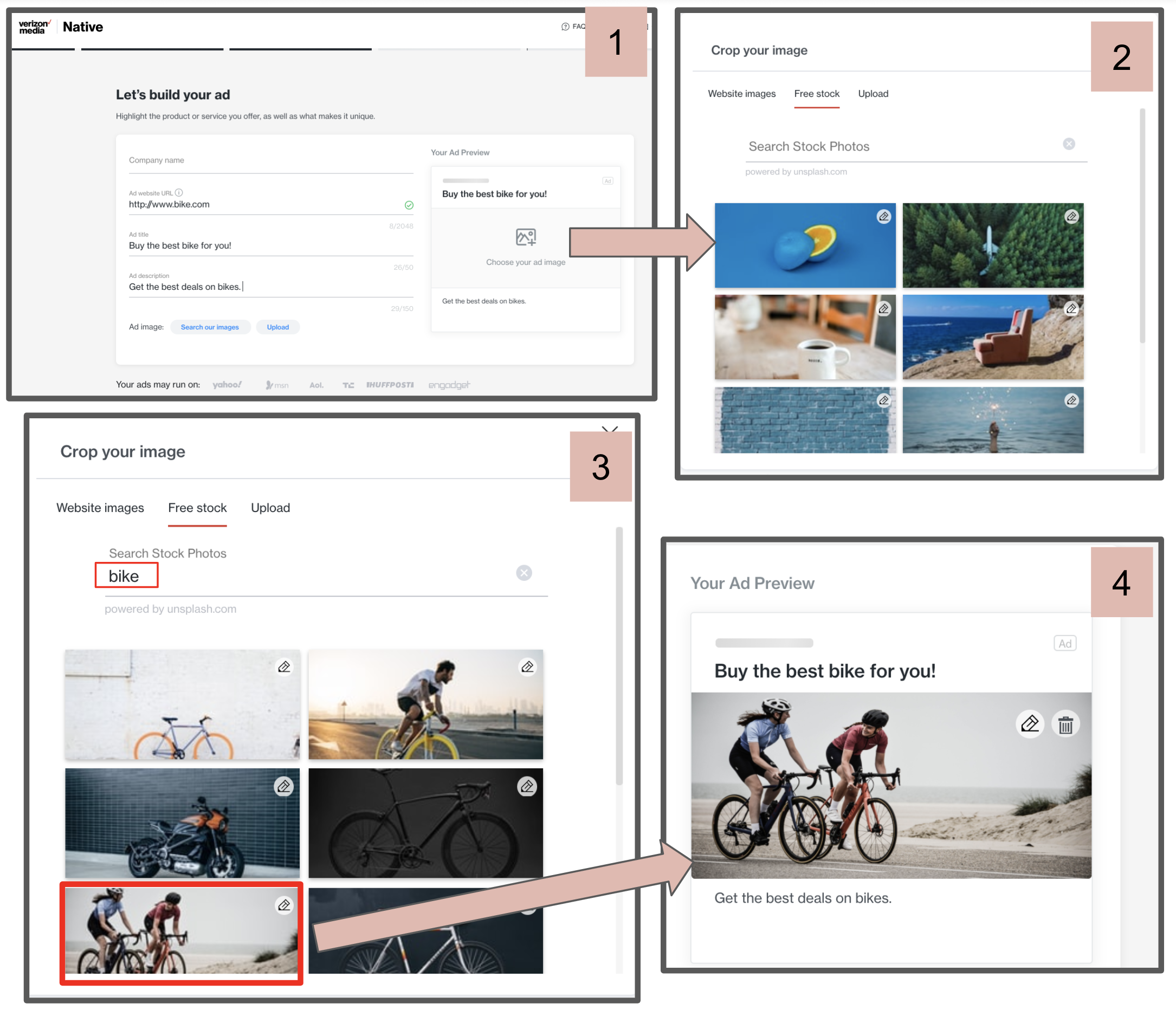}
  \caption{Onboarding workflow for Verizon Media Native advertisers: (1) advertiser enters their website URL with ad title and description (ad text), (2) advertiser proceeds to choose an ad image and \textit{may be} shown a random selection of images from a third party image library, (3) advertiser queries the library with a relevant textual query, and (4) selects a query result to get a preview of the final ad.}
  \label{fig:pull_figure}
\end{figure}

Online advertising is an effective tool for both major brands, and small businesses to grow awareness about their products and influence online users towards purchases \cite{mappi_CIKM,gemx_kdd}.
Ad images are a crucial component of online ads, and small businesses with limited advertising budget may find it challenging to obtain relevant high quality images for their ad campaigns.
In this context, small business are increasingly relying on online stock image libraries \cite{unsplash} which offer access to high quality and copyright free images available for marketing purposes.
To assist such small businesses in a self-serve manner, the Verizon Media Native (Yahoo Gemini) ad platform offers onboarding advertisers the ability to query a third party stock image library for candidate ad images (as shown in Figure~\ref{fig:pull_figure}).

Although the example (an ad for bikes) shown in Figure~\ref{fig:pull_figure} is borrowed from Verizon Media Native, it brings up a fundamental question: can we automatically figure out the search query for images given the ad text? Such automation will not only attract advertisers towards the stock image library, but will also speed up the onboarding process. However, there are several challenges in this context: (i) limited data (for supervised learning) on ad text and associated image search queries, and (ii) third party image libraries with unknown (proprietary) image indexing. The first challenge, \emph{i.e.}, limited data stems from the fact that this is a relatively new service (\emph{i.e.}, ad platforms offering stock image search feature), and small businesses may not be aware about such features. One can use current state-of-the-art image understanding methods (\emph{i.e.}, object detection \cite{openimages}, captioning \cite{sharma2018conceptual}) on the large set of existing ad images (which were not created by querying a stock image library) to get image annotations given ad text. But in our analysis (details in Section~\ref{sec:data}), such annotations were not specific enough to be considered as image queries in most cases, and hence the problem of limited ad text-to-image query data for supervised learning still remains. The second challenge listed above, \emph{i.e.}, unknown indexing, makes image libraries act like black boxes allowing short (few words) queries which may often be restricted to the English language. Due to such a black-box nature of third party image libraries, we do not consider approaches requiring the mapping of a text query to an image.

Given the above challenges, in this paper, we focus on unsupervised approaches for figuring out the ad image search query given an ad text. As we explain later in this paper (in Section~\ref{sec:data}), available logged data on ad image search behavior shows that in a significant fraction of cases, the ad text may already contain a keyword suitable as the image search query (as in the bike ad example in Figure~\ref{fig:pull_figure}); in other words the image query is \textit{extractive} in nature. In the remaining cases, the ideal image search query may be symbolic (abstractive) with respect to the input ad text. Citing a real example from logged data, the (anonymized) ad `Best ETFs to buy $\ldots$' had image search queries: 'resort' and `private jet', both of which are indicative (symbolic) of wealth after plausibly investing in the advertised financial products. Although we mostly focus on methods for extractive queries in this paper, we report preliminary findings on the usage of symbolic image queries given ad text.

\begin{figure}[]
\centering
  \includegraphics[width=0.8 \columnwidth]{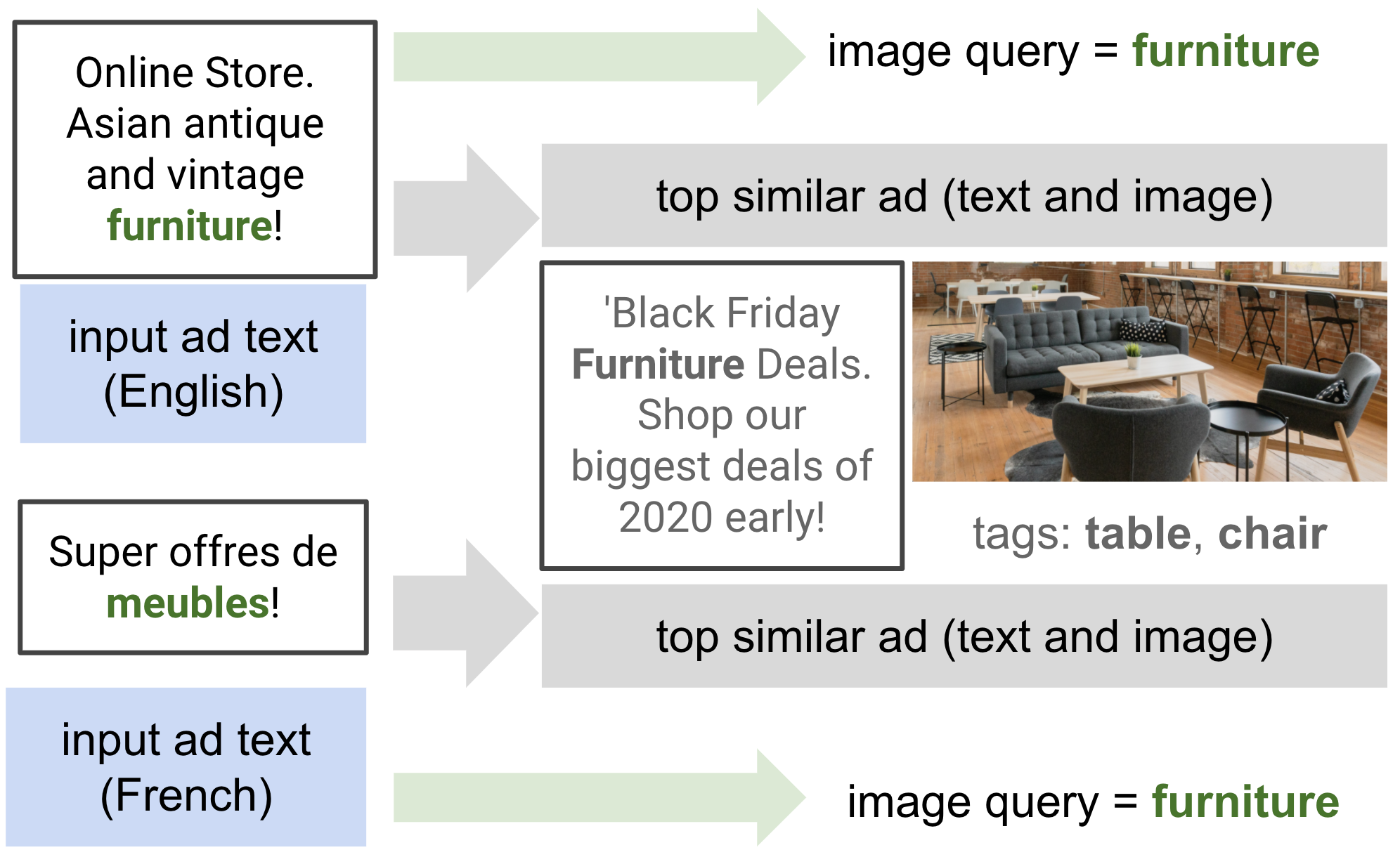}
  \caption{Illustrative example where an existing similar ad has image query \textit{hints} about two input ads (one in English and one in French) advertising furniture. The image tags are objects detected in the raw ad image but may not be descriptive enough to serve as image queries in many cases.}
  \label{fig:intro_idea}
\end{figure}

Focusing on extractive image queries, we build on prior work on unsupervised graph based keyword extraction \cite{biased_textrank, pke} (none of which are specifically designed for ad text). 
Such graph based methods typically build a graph where nodes are tokens (\emph{e.g.}, words) from input text, and the weighted edges between the nodes represent similarity between the associated tokens \cite{biased_textrank}.
With such a graph built from tokens, PageRank style algorithms \cite{textrank} are used to infer the dominant token (keyword); the popularity of such methods spans more than a decade. Recently, biased TextRank \cite{biased_textrank} was proposed to leverage sentence-BERT (SBERT) \cite{sbert_paper} embeddings of the input text to adjust the bias for each node (random restart probability in Page Rank). In other words, tokens which are closer to the overall \textit{meaning} of the entire text (as encoded by SBERT) are likely to have higher bias; this works significantly better than TextRank.
In our setup focused on ad text (and ad image queries),
we improve upon biased TextRank using two key ideas.
Firstly, we retrieve similar existing ads to augment the (relatively short) input ad text. The augmentation is done using the text of similar ads as well as image tags (detected objects) in the images of similar ads in an unsupervised manner.
Figure~\ref{fig:intro_idea} illustrates this idea: for both the English and French ad text in the example, a semantically similar English ad already contains hints (in the image and text), that the original ad is about furniture (which is also the ground truth image query in the example for both ads).
The above ideas not only make VisualTextRank ads specific, but also significantly push performance on the ad text-to-image query task.
Our main contributions can be summarized as follows.
\paragraph*{(1) Keyword extraction for ad text-to-image query}
We formulate ad text-to-image query as a keyword extraction problem.
For this task,
we propose an unsupervised graph based content extraction method (VisualTextRank) which builds on biased TextRank \cite{biased_textrank} by introducing category biasing, and input text augmentation using the text and images of similar (existing) ads. For our task, we obtain $\sim 11\%$ lift in offline accuracy using VisualTextRank compared to competitive keyword extraction baselines.

\paragraph*{(2) Cross-lingual learning}
Leveraging VisualTextRank, we extend the ad-text-to-image task for non-English ad text (including German, French, Spanish, and Portuguese)
and English-only image queries. For such a setup, we demonstrate the benefit of using semantically similar English ads to augment the non-English ad
text for the ad-text-to-image query task.

\paragraph*{(3) Product impact} We productionized a light-weight version of VisualTextRank for online tests with Verizon Media Native advertisers qualifying as small businesses (\emph{e.g.}, with relatively low advertising budgets).
For advertisers with English ad text, we compared the effect of automatically showing ad text-relevant images (via VisualTextRank queries) as initial (default) images to the advertiser versus showing random stock images. Against this baseline, 
we observed a 28.7\% lift in the rate of advertisers selecting stock images, and a 41.6\% lift in the onboarding completion rate for advertisers.

The remainder of the paper is organized as follows. Section~\ref{sec:related} covers related work, and Section~\ref{sec:data} covers data analysis. 
The proposed VisualTextRank method and its multilingual extension is covered in Sections \ref{sec:method} and \ref{sec:multilingual}.
Experimental results are covered in Section~\ref{sec:results}, and we end with paper with a discussion in Section~\ref{sec:discussion}.

%% file: related.tex
\section{Related Work} \label{sec:related}

\subsection{Keyword extraction}
In this paper, we formulate the ad text-to-image query task as an unsupervised keyword extraction problem (explained in Section~\ref{sec:method}).
Unsupervised keyword extraction research has a long history spanning over a decade \cite{pke}. In this paper, we focus on graph based methods which are inspired from PageRank on a graph created from the input text \cite{textrank, pke}. TextRank is a widely used method in this class, with applications spanning keyword extraction, summarization, and opinion mining \cite{biased_textrank, textrank}.
A notable recent work is by \cite{biased_textrank}, where the authors improve TextRank by introducing sentence-BERT \cite{sbert_paper} based biasing leading to the biased TextRank method.
Sentence-BERT (SBERT) \cite{sbert_paper} is a sentence representation method based on BERT \cite{devlin2018bert}.
In particular, SBERT is a modification of pretrained BERT that uses siamese and triplet network structures to derive semantically meaningful sentence embeddings which can be compared using cosine-similarity.
This provides a computationally efficient way to achieve
state-of-the-art results for sentence-pair regression tasks
like semantic textual similarity.
Our proposed VisualTextRank method builds on top of biased TextRank \cite{biased_textrank} with a focus on ad text, and the ad text-to-image query task. 

\subsection{Textual description of images: object detection and image captioning}
Object detection methods (image tagging) predict the category and location details of the objects present in an image. Faster R-CNN \cite{ren2015faster} is a popular method for this task, and more recent works built on top of it
\cite{qiao2020detectors}.
Compared to image tagging which provides a bag-of-words style textual description (objects tags with confidence scores) of an image, captioning models generate a natural language description; this is a much harder task than object detection. Current state-of-the-art image captioning models use a pre-trained object detector to generate features and spatial information of the objects present in an image (\emph{e.g.}, as in Oscar \cite{li2020oscar}). In particular,
\cite{li2020oscar} utilizes BERT-like objectives to learn cross-modal representation on different vision-language tasks (similar ideas form the basis of recent pre-trained multi-modal models, \emph{e.g.}, VisualBERT \cite{visualBERT_acl}). Prior captioning approaches have involved attention mechanisms 
and their variants to capture spatial relationship between objects \cite{herdade2019image} for generating captions. In this paper, we explore both object detection (image tagging) and captioning as features for assisting the ad text-to-image query task. In addition, since we focus on an unsupervised approach, we do not consider the possibility of fine-tuning the above mentioned pre-trained multi-modal models for our task.
\subsection{Ad image and text understanding}
Studying ad images and text using state-of-the-art deep learning models in computer vision and natural language processing (NLP) is an emerging area of research.
In \cite{cvpr_kovashka}, ad image content was studied using computer vision models, and their dataset had manual annotations for: ad category, reasons to buy products advertised in the ad, and expected user response given the ad.
Using this dataset, \cite{self_recsys2019, www20_joey} used ranking models to recommend themes for ad creative design using a brand's Wikipedia page. In \cite{cikm2020_createbetterads}, object tag recommendations for improving an ad image was studied using data from A/B tests. Although related to ads, the above methods are not applicable in our setup due to the lack of sufficient (ad text-to-image query) data for supervised training.



%% file: data.tex
\section{Data Insights} \label{sec:data}
In this section, we cover data insights around ad image search behavior which guided our proposed approach for the ad text-to-image task.
We first explain the data source for the analysis, followed by metadata from image understanding models (for the purpose of analysis), and the resultant insights.

\subsection{Data source}
To gather preliminary insights around ad text-to-image search behavior,
we collected data from a sample of $\sim 300$
advertisers who used the stock image library feature for their ad image while launching ad campaigns (onboarding) in the Verizon Media Native (Yahoo Gemini) ad platform. For each advertiser, the data included: (i) ad text, (ii) raw ad image, and (iii) image query for the ad image.
We will refer to this dataset as onboarded-sample.

\subsection{Image tags and caption metadata}
For the purposes of analyzing onboarded-sample for insights,
we used image tags, and captions as described below. We obtained image tags using the pre-trained Inception Resnet v2 object detection model \cite{openimages}.
The model consists of about $5,000$ classes (possible tags). For each image the model returns a list of inferred tags with confidence scores (we used tags with confidence above $0.8$).
For captioning, we used an object relation transformer (ORT) model \cite{herdade2019image} with a Faster R-CNN object detector.
For our analysis, we trained two ORT models: one on Microsoft COCO 2014 Captions dataset \cite{lin20214microsoft} (COCO-captions model), and the other on Conceptual Captions dataset \cite{sharma2018conceptual} (CC-captions model). 

\subsection{Insights}
\subsubsection{Query length, parts-of-speech, and ad text overlap}\label{sec:pos_insight}
We found that $81\%$ of image queries were a single word, and the remaining $19\%$ of queries comprised of two or more words.
For English ads in onboarded-sample, we found an overwhelming $92\%$ of queries to be either nouns or proper nouns; verbs accounted for $3\%$. The parts-of-speech (POS) tags were inferred via Spacy \cite{spacy}. Finally, we found that $60\%$ of image queries were already present (as a word or as a phrase) in the ad text. This indicates majority of the image queries are extractive in nature; $~40\%$ of remaining queries are symbolic in nature (\emph{e.g.}, an ad about retirement investments with image queries like `vacation' and `resort' which are not part of the ad text).

Supported by the above observations on query length and overlap with ad text, we formulate the ad text-to-image task as a keyword extraction task given ad text (and additional side information as explained later in Section~\ref{sec:problem_formulation}). For our experiments, we restrict the output to be either a noun, proper noun or verb (guided by the POS tags observation above).

\subsubsection{Poor correlation with image captions (but better with image tags)}
\label{sec:image_tags_captions}
\begin{figure}[]
\centering
  \includegraphics[width=0.9 \columnwidth]{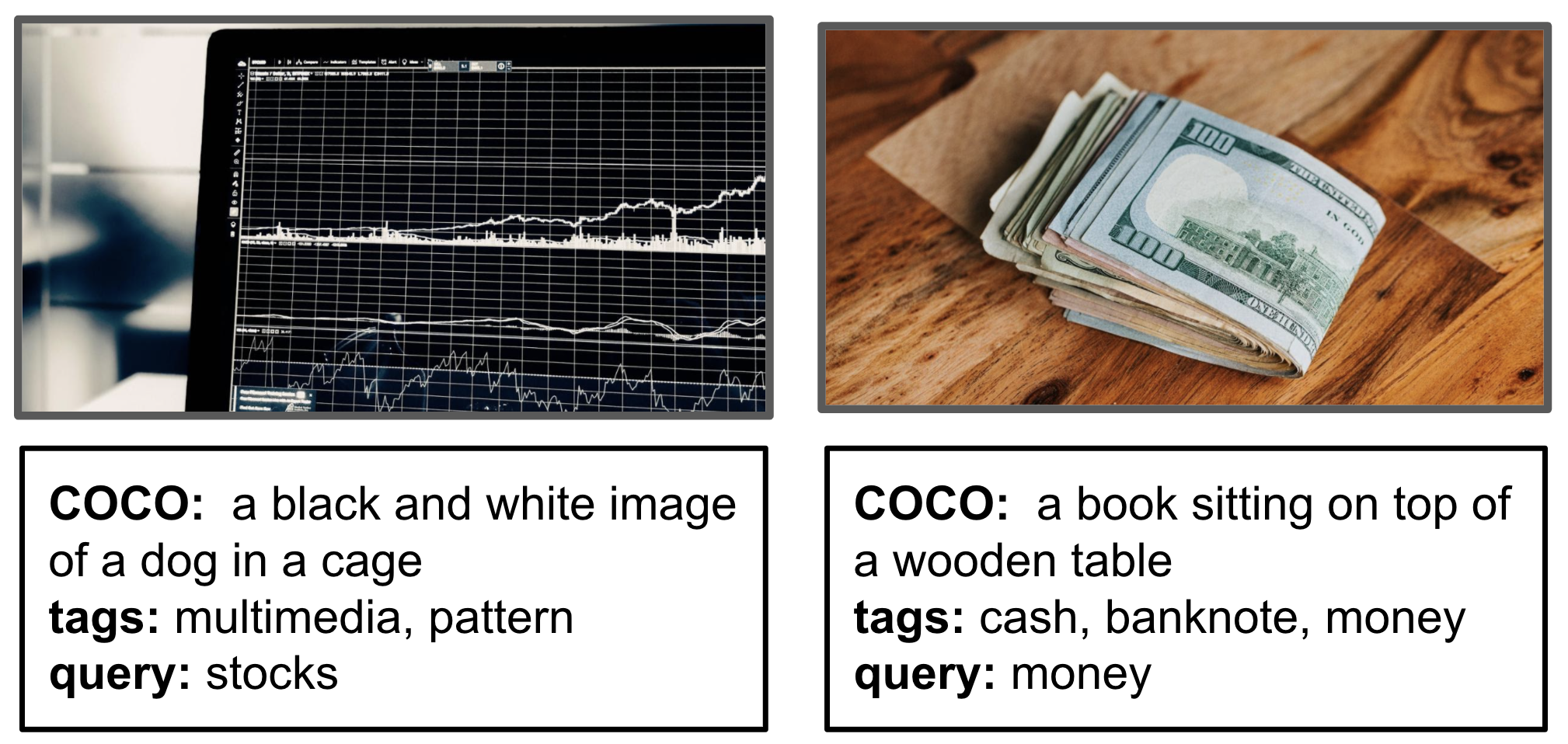}
  \caption{Examples of ad images with captions (COCO), tags, and ground truth image query.}
  \label{fig:bad_caption}
\end{figure}

The idea of using image understanding models stems from the observation that there is a huge dataset of existing ad text and images which were not created via a stock image query (\emph{e.g.}, the public dataset of older ads in \cite{kovashka_eccv2018}, or proprietary older ads in the ad platform). If it is possible to get descriptions for existing ad images via an image understanding tool (\emph{e.g.}, captions \cite{sharma2018conceptual}, object detectors \cite{openimages}), one can treat such a description as a proxy image query, and train an end-to-end model for ad text-to-image query. To test this hypothesis, we analyzed the captions and image tags of onboarded ad images and compared them with the ground truth image queries.
Checking for image query overlap with captions for the raw ad image, we found that there was an overlap in $5\%$ of samples for the COCO-captions model and  an overlap in $7.6\%$ of samples for the CC-captions model.
Figure~\ref{fig:bad_caption} shows two examples of ad images from the onboarded sample with captions (COCO), tags and the ground truth query; clearly, captioning failed in both examples, while tagging did better in one.
In comparison to captions, for the image tag model the overlap with image queries was $17\%$. Due to higher overlap, we retained image tags for existing ads (used in our experiments), and did not focus on captions. Note that with $17\%$ overlap, image tags are still not expressive enough (just $5,000$ labels) to create a labeled dataset for training an ad-text-to-image query predictor; but they can plausibly be used as a signal for better keyword extraction. Table~\ref{tab:data_insights} provides a brief summary of the insights.
\begin{table}[h]
    \centering
    \begin{tabular}{|l|c|c|}
    \hline 
feature                   & image query insights      \\
\hline	
\hline
POS tags                   &     $92\%$ NOUN/PROPN, $3\%$ VERB                                      \\
\hline
query length      	       & $81\%$ of queries are a single word	  \\
\hline
ad text          	       & $60\%$ of queries are within the ad text           \\
\hline
image captions (COCO)      	   & 5\% query overlap	         		\\
\hline
image captions (CC)      	   & 7.6\% query overlap	         		\\
\hline
image tags      	       & $17\%$ query overlap \\
\hline
\end{tabular}
    \caption{Summary of insights from onboarded-sample data.}
    \label{tab:data_insights}
\end{table}

%% file: method.tex
\section{Keyword Extraction for Ad Text-to-Image Query} \label{sec:method}
In this section, we first formulate ad text-to-image query as a keyword extraction task in Section~\ref{sec:problem_formulation}.
Next, we describe the TextRank method as necessary background in Section~\ref{sec:textrank}. Finally, we explain our proposed method VisualTextRank in Section~\ref{sec:visualtextrank}.

\subsection{Problem Formulation} \label{sec:problem_formulation}
Given an ad text $\mathbf{a}_{in}$, our objective is to come up with a single keyword to serve as the ad image query (to a stock image library).
We assume the presence of a pool of existing ads $\mathbf{P}$, where each ad $\mathbf{a}_i \in \mathbf{P}$ has the following attributes: (i) ad text, and (ii) image tags in the raw ad image (as described in Section~\ref{sec:image_tags_captions}).

In this paper, we focus on a graph based unsupervised method to model the ad text (and information relevant to $\mathbf{a}_{in}$ in $\mathbf{P}$), for the purpose of extracting a word (from the union of $\mathbf{a}$ and $\mathbf{P}$) as a relevant ad image query given $\mathbf{a}_{in}$.
The key motivation behind a graph based approach as opposed to just using the image tags from the most similar ad text (\emph{i.e.}, a nearest neighbor approach), lies in Table~\ref{tab:data_insights}. The intuition here is to identify the \textit{central} entity in the ad text (which contains the query in $60\%$ of cases as noted in Table~\ref{tab:data_insights}), with additional help from similar ads.
In the following sections, we first explain an existing graph based keyword extraction method (TextRank) in our context as necessary background, and then explain the proposed VisualTextRank method.

\subsection{TextRank} \label{sec:textrank}
At a high level, TextRank is primarily based on PageRank \cite{textrank} on the graph of tokens (words) obtained from the input text. We explain below, the token graph construction, the (original) unbiased version of TextRank as well as the recently proposed biased version.

\paragraph*{Token graph} The token-graph $\mathcal{G} = (\mathbf{V}, \mathbf{E})$ is formed from tokens (words) from the input ad text $\mathbf{a}_{in}$, where $\mathbf{V}$ denotes the set of vertices and $\mathbf{E}$ denotes the set of (undirected) edges. Each word $w_i$ is mapped to a vertex $v_i$. An edge between two vertices has weight:
\begin{align}
e_{ij} = < \phi(w_i), \phi(w_j) >.
\end{align}
where $ < \phi(w_i), \phi(w_j) >$
denotes the cosine similarity between the (word) embeddings of $w_i$ and $w_j$ (denoted by $\phi(w_i)$ and $\phi(w_j)$ respectively). It is common to set $e_{ij}$ to zero if it is below a similarity threshold \cite{biased_textrank}. 

\paragraph*{Unbiased TextRank}
The original (unbiased) TextRank method \cite{textrank} iteratively computes the score $v_i$ of each vertex $i \in \mathbf{V}$ as:
\begin{align} \label{eq:unbiased_textrank}
v_i = (1-d) + d \times \left( \sum_{j \in \mathcal{N}(i)} \frac{e_{ij}}{\sum_{k \in \mathcal{N}(j)} e_{kj}} \times v_j  \right),
\end{align}
where $d$ is the damping factor (typically $0.85$), and $\mathcal{N}(i)$ is the set of vertices which share an edge (with non-zero edge weight) with vertex $i$.

\paragraph*{Biased TextRank}
The recently proposed biased TextRank method \cite{biased_textrank} iteratively computes the score $v_i$ of each vertex in the token-graph in the following manner:
\begin{align} \label{eq:biased_textrank}
v_i = bias_i \times (1-d) + d \times \left( \sum_{j \in \mathcal{N}(i)} \frac{e_{ij}}{\sum_{k \in \mathcal{N}(j)} e_{kj}} \times v_j  \right),
\end{align}
where the only change with respect to the unbiased version in \eqref{eq:unbiased_textrank} is the addition of a bias term $bias_i$ for vertex $i$. For the task of keyword extraction, in \cite{biased_textrank} bias $bias_i$ is defined as:
\begin{align} \label{eq:self_bias}
bias_i = < \phi(\mathbf{a}_{in}) , \phi(w_i) >   ,
\end{align}
where $\phi(\mathbf{a}_{in})$ is the sentence embedding for input ad text $\mathbf{a}_{in}$.
Intuitively, biased TextRank tries to favor words which are closer to the overall (semantic) meaning of the input ad text.
In \cite{biased_textrank}, the authors demonstrate that this intuition works well and show the superiority of biased TextRank over the original TextRank method.
In the remainder of the paper, we will refer to the above biasing method (as in~\eqref{eq:self_bias}) as self-biasing since it computes the bias with respect to the input (ad text). Sentence-BERT (SBERT) \cite{sbert_paper} is an effective sentence embedding method used in biased TextRank \cite{biased_textrank}, and we leverage the same in our proposed method (explained below).

\subsection{VisualTextRank} \label{sec:visualtextrank}
We start with an overview of the proposed VisualTextRank method in Section~\ref{sec:visualtextrank_overview}, and then go over the details of important components in Sections~\ref{sec:category_biasing}
and~\ref{sec:graph_augmentation}.

\subsubsection{Overview} \label{sec:visualtextrank_overview}
The biased TextRank method~\eqref{eq:biased_textrank} has the following shortcomings when it comes to understanding an ad (\emph{i.e.}, extracting a keyword for image search):
\begin{enumerate}
    \item ads are usually tied to categories (\emph{e.g.}, IAB category taxonomy \cite{iab_mopub}) and TextRank is oblivious to this, and
    \item ad text can be short leading to a very sparse token graph with no or negligible edges above a reasonable similarity threshold.
\end{enumerate}
In addition to the above shortcomings from a keyword extraction perspective, TextRank's original motivation was not to extract queries suitable for ad images, and hence it lacks the visual aspect needed in an image search query (in both extractive and symbolic cases).
In VisualTextRank, we build on the above shortcomings as outlined below (detailed explanation after the outline).
\begin{enumerate}
    \item Category biasing: we introduce an ad category specific biasing term in addition to the self-biasing term in \eqref{eq:self_bias}. 
    \item Augmentation with similar ads' text: for a given input ad, we fetch semantically similar ads from a pool of existing ads. We augment the token-graph with a filtered version of the text from similar ads.
    \item Augmentation with similar ads' image tags: we also augment the token-graph with a filtered version of the image tags (as described in Section~\ref{sec:image_tags_captions}) of similar ads' images. 
\end{enumerate}
Steps (2) and (3) above, not only alleviate the problem of short ad text, but also offer the capability of going beyond words in the ad text while coming up with the image query. As a result of augmenting the initial token-graph $\mathcal{G}$ (as described in Section~\ref{sec:textrank}) with text and image tags from similar ads (augmentation details described later in Section~\ref{sec:graph_augmentation}), we obtain an augmented token-graph $\mathcal{G}^{*} = (\mathbf{V}^{*}, \mathbf{E}^{*})$.

Combining the steps above, the vertex value update for VisualTextRank can be written as follows.
\begin{align} \label{eq:visualtextrank}
v_i & = bias^{(self)}_i \times bias^{(cat)}_i \times (1-d) \nonumber \\  & \quad \quad + d \times \left( \sum_{j \in \mathcal{N}^{*}(i)} \frac{e_{ij}}{\sum_{k \in \mathcal{N}^{*}(j)} e_{kj}} \times v_j  \right),
\end{align}
where $i$ is a vertex in the augmented graph $\mathcal{G}^{*}$, and $\mathcal{N}^{*}(i)$ denotes the set of vertices $i$ shares an edge with in $\mathcal{G}^{*}$. The details behind the category biasing term $ bias^{(cat)}_i$ are described in Section~\ref{sec:category_biasing}, and the details behind the graph augmentation leading to $\mathcal{G}^{*}$ are described in \ref{sec:graph_augmentation}.

\subsubsection{Token-graph augmentation with similar ads}
\label{sec:graph_augmentation}
Token graph augmentation has three steps as described below.

\paragraph*{Retrieving similar ads}
For augmentation, we assume a pool (set) of existing ads denoted by $\mathbf{P}$.
An ad in $\mathbf{P}$ has the following attributes: (i) ad text, and (ii) image tags for the ad image along with confidence scores for each tag.
For each ad in the pool $\mathbf{P}$, we compute the semantic similarity (relevance) of the ad with respect to the input ad $\mathbf{a}_{in}$ as follows:
\begin{align}
relevance(\mathbf{a}, \mathbf{a}_{in}) = < \phi(\mathbf{a}), \phi(\mathbf{a}_{in} ) >,
\end{align}
where $\mathbf{a}$ is an ad in $\mathbf{P}$, and $relevance(\mathbf{a}, \mathbf{a}_{in})$ denotes its relevance with respect to the input ad $\mathbf{a}_{in}$. A sentence embedding method like SBERT \cite{sbert_paper} (trained for semantic textual similarity) can be used to obtain $\phi(\mathbf{a})$ and $\phi(\mathbf{a}_{in})$. We use the above relevance score to obtain the top-$m$ similar ads from the pool $\mathbf{P}$ for the given input ad $\mathbf{a}_{in}$.

\begin{figure}[]
\centering
  \includegraphics[width=1 \columnwidth]{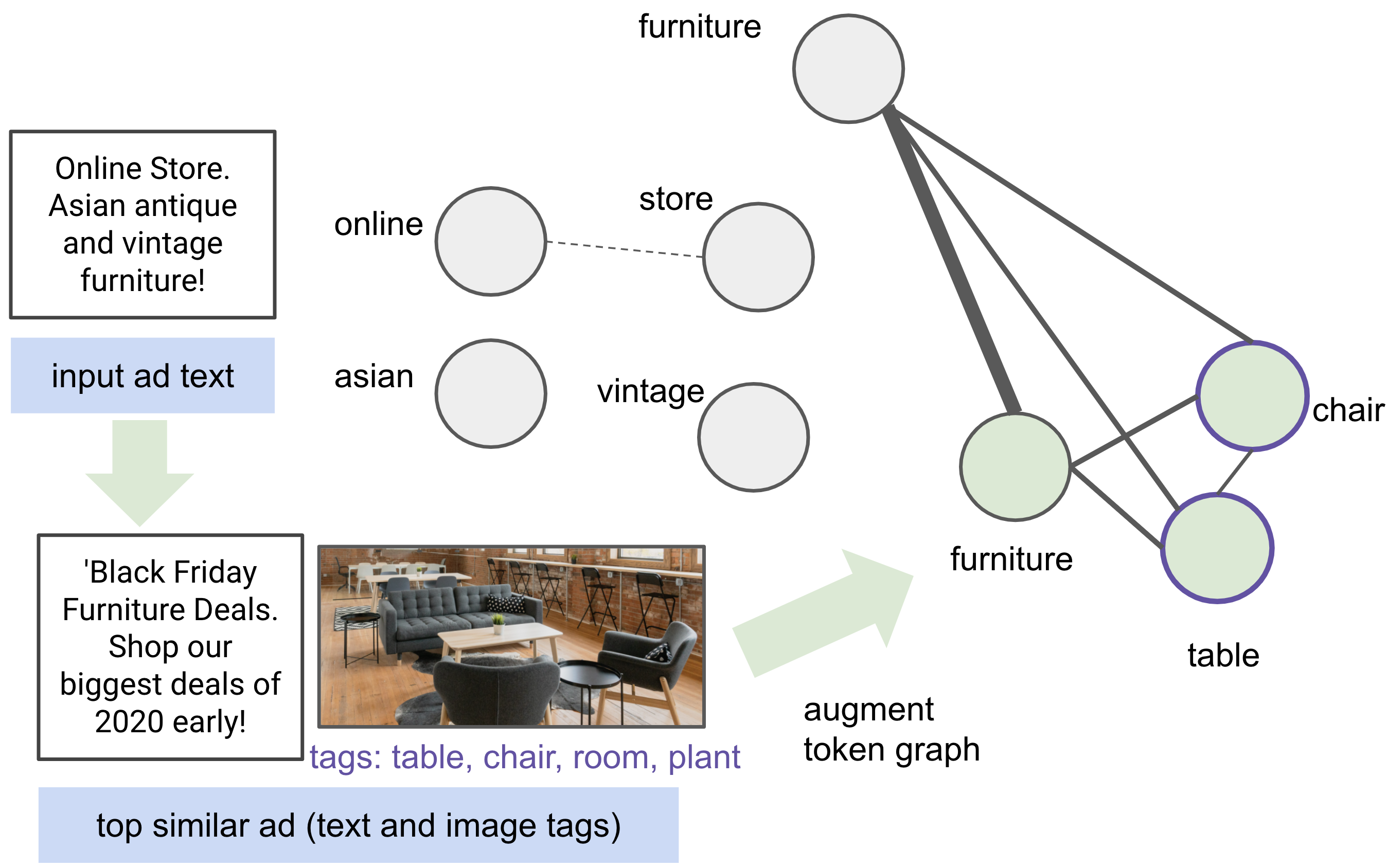}
  \caption{Illustrative example of VisualTextRank in action for a furniture ad. The green nodes in the token graph are augmented using the similar ad's text and image tag, and eventually lead to `furniture' being selected (as they form a distinct cluster with other furniture related words).}
  \label{fig:visualtextrank_example}
\end{figure}

\paragraph*{Augmentation with similar ads' image tags}
We augment the token-graph~$\mathcal{G}$ with the image tags of similar ads', by selecting image tags which are semantically close to a word in the ad text of the similar ad. For example, as shown in Figure~\ref{fig:visualtextrank_example}, if the similar ad has a word like `furniture', and the corresponding ad image has tags like `chair' and `table' (which are semantically close to furniture), we select such tags for augmenting the original token-graph $\mathcal{G}$.
The details of our proposed image tag augmentation can be described as follows.
We assume a bound on: (i) the maximum number of image tags that we can add ($=max\_tags$), and (ii) the minimum cosine similarity between an image tag and a word in its parent ad text to select the image tag for augmentation ($=min\_tag\_sim$).
As described in Algorithm~\ref{alg:top-k-tags}, we start with a list of similar ads $\{\mathbf{a}_i, \ldots, \mathbf{a}_m \}$ sorted in decreasing order of relevance to the input ad text $\mathbf{a}_{in}$. For each similar ad $\mathbf{a}_i$, $\mathcal{T}(\mathbf{a}_i)$ denotes the list of image tags associated with the corresponding ad image; this tags list is sorted in decreasing order by the confidence scores from the object detector.
Given $\mathbf{a}_i$ and $\mathcal{T}(\mathbf{a}_i)$, we select a tags $t \in \mathcal{T}(\mathbf{a}_i) $ which are close to at least one word $w$ in the $\mathbf{a}_i$ (\emph{i.e.}, $<\phi(t),\phi(w)>)\geq min\_tag\_sim$).
We keep iterating as shown in Algorithm~\ref{alg:top-k-tags}, till we collect $max\_tags$ images tags in $\mathcal{T}^*$.
\begin{algorithm}
\caption{Augmentation with similar ads' image tags.}
  \label{alg:top-k-tags}
 \KwData{similar ad texts~$\mathbf{a}_1,...,\mathbf{a}_m$ (sorted by decreasing relevance to input), image tags list $\mathcal{T}(\mathbf{a}_i)$  for ad $\mathbf{a}_i$} 
 \KwResult{image tags set~$\mathcal{T}^*$ to augment~$\mathcal{G}$, 
 $|\mathcal{T}^*|= max\_tags$}
 $\mathcal{T}^*=\emptyset$\;
 \For{$i \in \{1,...,m\}$}
 {\For{$t\in \mathcal{T}(\mathbf{a}_i)$}{
 \If{$\max\limits_{w\in \mathbf{a}_i}(<\phi(t),\phi(w)>)\geq min\_tag\_sim $} 
 {$\mathcal{T}^*=\mathcal{T}^*\cup t$\;
 \If{$|\mathcal{T}^*|= max\_tags$}{\Return {$\mathcal{T}^*$}}}}}
\end{algorithm}

\paragraph*{Augmentation with similar ads' text}
In addition to image tags, we also use words from similar ads to augment the token-graph~$\mathcal{G}$.
Our proposed word augmentation method is described in Algorithm~\ref{alg:top-k-words}.
It is similar in spirit to the image tag augmentation method. We keep iterating over words in an ad text $\mathbf{a_i}$ in their order of occurrence, and select words which are above the $min\_word\_sim$ similarity threshold (cosine similarity between the input ad text $\mathbf{a}_{in}$ and the candidate word $w$ using SBERT embeddings). The set of selected words $\mathcal{W}^*$ is used to augment $\mathcal{G}$.
\begin{algorithm}

 \KwData{Similar ad texts~$\mathbf{a}_1,...,\mathbf{a}_m$ (sorted by decreasing relevance to input)}
 \KwResult{words set~$\mathcal{W}^*$ to augment~$\mathcal{G}$, namely ~$\mathcal{G}^*=\mathcal{G}\cup \mathcal{W}^*, |\mathcal{W}^*|=max\_words$}
 $\mathcal{W}^*=\emptyset$\;
 \For{$i\in \{1,...,m\}$}
 {\For{$w\in$ $\mathbf{a}_i$}{
 \If {$(<\phi(\mathbf{a}_{in}),\phi(w)>) \geq min\_word\_sim$}
 {$\mathcal{W}^*=\mathcal{W}^*\cup w$\;
 \If{$|\mathcal{W}^*|= max\_words$}{\Return {$\mathcal{W}^*$}}}}}
 \caption{Augmentation with with similar ads' text.}
  \label{alg:top-k-words}
\end{algorithm}
In the example in Figure~\ref{fig:visualtextrank_example},
the word `furniture' from the similar ad's text was selected for augmentation using the above method. 


\subsubsection{Category biasing} \label{sec:category_biasing}
We consider a set of \textit{category} phrases $\mathbf{C}$. Such a set can be created by using phrases from a standard (flattened) ad category taxonomy, \emph{e.g.}, IAB categories \cite{iab_mopub} (see Appendix for details). For example, the set $\mathbf{C}$ can contain words and phrases like `travel', `legal services', and `fashion' which are used to denote different categories in the IAB taxonomy.
Given such a set, we intend to find the closest category $\hat{c}$ in the following manner
\begin{align} \label{eq:max_cat_sbert}
\hat{c} = \argmax_{c \in \mathbf{C} }{ < \phi(c) , \phi(\mathbf{a}_{in}) > }  
\end{align}
where $\phi(c)$ and $\phi(\mathbf{a}_{in})$ denote (SBERT) embeddings for the category phrase and the input ad text.
We do not assume the presence of category labels for a given ad text in our setup, and the above method is a proxy to infer the closest category for a given ad text.
Having inferred $\hat{c}$ for given input ad text,
we compute the category bias $bias_i^{(cat)}$ for each vertex $i$ in the augmented graph $\mathcal{G}^{*}$ as:
\begin{align}
  bias_i^{(cat)} = < \phi(\hat{c}) , \phi(w_i)  >,
\end{align}
where $w_i$ denotes the word associated with vertex $i$.
Intuitively, category biasing tries to up-weight the words closer to the inferred category. In the final vertex value update in \eqref{eq:visualtextrank}, $bias_i^{(cat)}$ is used in conjunction with the self bias as defined as in \eqref{eq:self_bias}.



%% file: multilingual.tex
\section{Multilingual Ad Text}
\label{sec:multilingual}
In the multilingual setting, we assume the following: (i) ad text can be non-English, (ii) the image query is in English, and (iii) we can retrieve semantically similar (existing) English ads.
Furthermore, in this paper, we assume access to a generic translation tool which can translate non-English ad text to English.
Given the above assumptions, for multilingual ad text, we simply translate the ad text from its original language to English, then use the English version with VisualTextRank (\emph{i.e.}, fetching similar English ads, and doing category biasing).
The intuition behind such an approach is that regardless of the language, visual components in the ad image will be strongly tied to the product mentions in the ad text. For example, for a furniture ad in English or an equivalent ad in French, the ad image is likely to have furniture.
By borrowing from text and image tags of semantically similar English ads for a given multilingual ad, VisualTextRank facilitates a simple yet effective form of cross lingual learning. It also does not rely on having a large set of existing ads in the ad's original language (except in the case of English ads).

In our experiments, we also explored the use of 
multilingual (xlm) SBERT \cite{sbert_paper} to: (i) fetch similar English ads for a given multilingual ad (without translating it), (ii) run VisualTextRank with such xlm-SBERT embeddings to get a keyword (translated to English by using word-level-translation). As we explain in Section~\ref{sec:non-ENG-eval}, this approach was inferior to the above ad text translation approach. 

%% file: results.tex
\section{Results} \label{sec:results}

\subsection{Offline evaluation dataset} \label{sec:eval_dataset}
We collected ad text to image query data from Verizon Media Native's stock image search feature (as illustrated in Figure~\ref{fig:pull_figure}). This feature is currently available in the onboarding workflow for small and medium businesses with relatively low advertising budgets.
For a given ad text, the advertiser might make multiple search queries before converging on the final choice for onboarding (launching the ad campaign). We consider all such queries as relevant (golden set) for the given ad text.
For example, for the ad text `Bank Foreclosing? Stop Foreclosure Now!', if the user searches for `stress', `eviction', and 'bills', all the three queries are labeled relevant queries.
We sampled the collected data to create an evaluation data set with $\sim 600$ ads with ad text in English, and $400$ non-English (multilingual) ads spanning German, French, Spanish, Portuguese, Swedish, and Italian ad text. For the set of English ads, the image queries were editorially reviewed and corrected for spelling mistakes, and we will refer to this set as the ENG-eval set.
For the multilingual set, each non-English ad was translated to English using an internal translation tool. The image queries were also editorially reviewed, translated to English (if they were not in English), and corrected for spelling mistakes as well (if they were already in English). We will refer to this as the non-ENG-eval set.
While processing raw data to create the non-ENG-eval set we observed a significant number of cases where the advertiser tried only non-English queries, or had spelling mistakes in the English query 
(plausibly due to lack of English proficiency). Hence, editorial review was needed to correct such samples for proper evaluation with our proposed methods (and baselines). Note that both ENG-eval and non-ENG-eval datasets are of a unique nature due to the integration of the stock image search feature in the ad platform (allowing us to map ad text to corresponding ad image search queries).

\subsection{Ads pool for similar ads retrieval} \label{sec:similar_ads_pool}
Since VisualTextRank leverages semantically similar ads
for augmenting the input ad text, we need a pool of ads from which we can retrieve similar ads given an input ad text. For our experiments, we collected a sample of $20,000$ English ads (US-only) from Verizon Media Native (Yahoo Gemini), with their ad text and the raw ad image. We obtained image tags for the raw image (as explained in Section~\ref{sec:image_tags_captions}); this set had $20,000$ ads which were not created using the stock image search feature, and hence did not have associated image queries.
The time range for the data pull was same as that for the evaluation datasets described in Section~\ref{sec:eval_dataset}.
In the remainder of this paper, we will refer to this set of existing ads as ads-pool.

\subsection{Evaluation metrics} \label{sec:metrics}
\paragraph*{Offline metrics} \label{sec:offline_metrics}
For offline evaluation on the ENG-eval and non-ENG-eval sets,
we consider three metrics as defined below. For a given ad text, if the predicted keyword is $w^{*}$, and the golden set of queries is the set $\mathbf{Q} = \{q_1, q_2, \ldots \}$, 
\begin{align}
\text{hard-accuracy}(w^*,\mathbf{Q}) 
               &= \begin{cases}
               1               & w^* \in \mathbf{Q} \\
               0 & \text{otherwise}
           \end{cases} \\
\text{soft-accuracy}(w^*,\mathbf{Q}) 
               &= \begin{cases}
               1               &  \max_{q_i \in \mathbf{Q}} <\phi(w^*), \phi(q_i)> \geq 0.8 \\
               0               & \text{otherwise}
           \end{cases} \\
\text{w2v-similarity}(w^*, \mathbf{Q}) &=  \max_{q_i \in \mathbf{Q}} <\phi(w^*), \phi(q_i)> 
\end{align}
where hard-accuracy simply checks whether the predicted keyword is in the golden set. Soft-accuracy uses the word2vec \cite{w2v} similarity (as implemented in Spacy \cite{spacy}, details in Appendix) between a query in the golden set and the predicted keyword (represented by $<\phi(w^*), \phi(q_i)>$) to if the predicted keyword is approximately (\emph{i.e.}, above the $0.8$ cosine similarity threshold) close to any of the golden set queries. This takes care of minor differences between the golden set queries and the predicted keyword (\emph{e.g.}, run and running would be considered similar). Finally, w2v-similarity is the absolute similarity between the predicted keyword and the closest query (in word2vec sense) in the golden set. Note that our accuracy metrics are essentially $precision@1$ since we focus on a single keyword as the ad image query.

\paragraph*{Online metrics} \label{sec:online_metrics}
An advertiser session is defined as a continuous stream of interactions of the advertiser with the onboarding UI, with gaps no more than $30$ minutes between consecutive events.
To gauge the effectiveness of automatic ad text-to-image search, we track two events in each advertiser session: (i) selecting (clicking) stock images to get a preview of the ad, and (ii) completing the onboarding process.
Tracking the image selection event, we define image \textit{selection rate} as:
\begin{align}\label{eq:image_click_rate}
    \text{selection rate} = \frac{\text{\# sessions with $\geq$ 1 image selections}}{\text{\# sessions}}.
\end{align}
Tracking the onboarding event, we define \textit{onboarding rate} as:
\begin{align}\label{eq:image_click_rate}
    \text{onboarding rate} = \frac{\text{\# sessions ended with onboarding}}{\text{\# sessions}}.
\end{align}
The online metrics listed above are expected to capture the effect of automatic image recommendations on the stock image selection rate and the onboarding completion rate. Note that an image selection event may or may not lead to an onboarding completion event (if the advertiser does not launch the ad campaign).

\subsection{Offline results}
We first describe multiple baseline methods for ad text-to-image query (most of them are existing keyword extraction methods), and then go over the offline results for VisualTextRank and the baselines (for both ENG-eval and non-ENG-eval sets).

\begin{table}[h]
    \centering
    \begin{tabular}{|l|c|c|c|c|}
    \hline 
model              & hard         & soft        & avg. w2v \\
                   & accuracy     & accuracy    & similarity     \\
\hline	
\hline
position rank	&	11.5\%	&	17.07\%	&	0.412	\\
\hline
YAKE	&	14.11\%	&	20.03\%	&	0.4655	\\
\hline
topical page rank	&	14.81\%	&	21.25\%	&	0.4515	\\
\hline
multipartite rank	&	15.51\%	&	21.78\%	&	0.4639	\\
\hline
topic rank 	&	16.2\%	&	23.17\%	&	0.4719	\\
\hline
tf-idf	&	18.12\%	&	23.69\%	&	0.486	\\
\hline
TextRank	            &	21.95\%     &  28.92\%   & 0.5297 \\
\hline
self-biased      	   &	         	&		    &		\\
TextRank               &     26.48\%  & 35.37\%	 &  0.5739 \\
\hline
VisualTextRank        & \textbf{28.05\%}	& \textbf{39.37\%}	 & \textbf{0.6147}                \\
\hline
\end{tabular}
    \caption{Offline metrics for ENG-eval.}
    \label{tab:baselines_eng}
\end{table}

\subsubsection{Baseline methods}\label{sec:baselines}
We use TF-IDF, existing graph based keyword extraction methods \cite{pke} (TextRank, Position Rank, Topic Rank, Topical Page Rank, Multipartite Rank, YAKE), as well as the recently proposed (self) biased TextRank \cite{biased_textrank} as baselines.
For all the baselines, as well as the VisualTextRank method, we consider only nouns, proper nouns and verbs as valid output keywords, \emph{i.e.}, we have a POS tag filter (as guided by the POS tag insight in Section~\ref{sec:data}). 

\subsubsection{ENG-eval results} \label{sec:ENG-eval}
Table~\ref{tab:baselines_eng} shows the performance of the proposed VisualTextRank method versus the baselines outlined in Section~\ref{sec:baselines} for the ENG-eval set.
As shown in Table~\ref{tab:baselines_eng}, VisualTextRank significantly outperforms the closest baseline (self) biased TextRank by $11.3\%$ in terms of soft-accuracy.

\subsubsection{Non-ENG-eval results} \label{sec:non-ENG-eval}
For brevity, Table~\ref{tab:baselines_non_eng} just shows the performance of the proposed VisualTextRank method versus the (self) biased TextRank method (best baseline in the ENG-eval case) as well as TextRank for the non-ENG-eval set; the results for other baselines as in Table~\ref{tab:baselines_non_eng}, are in line with the observations for the ENG-eval set.
\begin{table}[h]
    \centering
    \begin{tabular}{|l|c|c|c|c|}
    \hline 
model              & hard         & soft        & avg. w2v \\
                   & accuracy     & accuracy    & similarity     \\
\hline	
\hline
TextRank	           & 13.19\%  & 17.18\%  & 0.4403 \\
\hline
self-biased      	   &	         	&		    &		\\
TextRank              & 22.39\% &	29.75\%	 & 0.5302   \\
\hline
VisualTextRank       & \textbf{23.62\%}	  & \textbf{31.60\%}	  & \textbf{0.5563}                 \\
\hline
\end{tabular}
    \caption{Offline metrics for non-ENG-eval.}
    \label{tab:baselines_non_eng}
\end{table}
As shown in Table~\ref{tab:baselines_non_eng}, in terms of soft-accuracy, VisualTextRank outperforms (self) biased TextRank by $6.2\%$ in the multilingual setting as well. Note that VisualTextRank leverages similar English ads for multilingual (input) ad text in this setup (as explained in Section~\ref{sec:multilingual}).
Compared to the above result for VisualTextRank, the xlm-SBERT based approach described in Section~\ref{sec:multilingual} led to poorer metrics ($13.5\%$ soft accuracy, $0.404$ avg. w2v similarity).

\subsection{Ablation study}
In this section, we study the contribution of different components of VisualTextRank via offline performance on the ENG-eval set and non-ENG-eval set.
Table~\ref{tab:ablation} shows the impact of different components on offline metrics for the ENG-eval set.
Baseline result for the ablation study is a self-biased TextRank method (line 1 in Table~\ref{tab:ablation}), which achieves $35.37\%$ in terms of soft accuracy. With addition of category biasing to self biasing~(as explained in Section~\ref{sec:category_biasing}), we observe a soft-accuracy improvement to $37.8\%$ (line 3 in Table~\ref{tab:ablation}). The second major improvement ($39.02\%$, line 5 in Table~\ref{tab:ablation}) is from adding one image tag from similar English ads (as explained in Section~\ref{sec:graph_augmentation}). Putting it with similar ads' text (as explained in Section~\ref{sec:graph_augmentation}), we get the best result of $39.37\%$ soft accuracy (line 7 in Table~\ref{tab:ablation}). We notice that: (i) increasing number of tags or words from the similar ad does not help, because it takes away the attention from the ad text words (which are more likely to be in the golden set); 
(ii) adding image tags without pre-filtering by Algorithm~\ref{alg:top-k-tags} (line 6) gets lower soft accuracy than with the pre-filtering (line 5).

\begin{table}[h]
    \centering
    \begin{tabular}{|l|c|c|c|c|}
    \hline 
model              & hard         & soft        & avg. w2v \\
                   & accuracy     & accuracy    & similarity     \\
\hline	
\hline
self biased         &              &             &                 \\
TextRank            &     26.48\%  & 35.37\%	 &  0.5739 \\
\hline
cat biased         &              &             &                 \\
TextRank          &     27.70\%  &  35.71\%    &   0.5975                 \\
\hline
self+cat biased      &              &             &                 \\
TextRank        &     28.92\%	& 37.80\%	 & 0.6061          \\
\hline
self+cat biased  &              &             &                 \\
+ $1$-nbr text      &  28.57\%	& 38.15\% & 0.6074              \\
\hline
self+cat biased   &              &             &                 \\
+ $1$-nbr img tag     &  28.75\%	& 39.02\%	 & 0.611              \\
\hline
self+cat biased   &              &             &                 \\
+ $1$-nbr img tag     &  28.92\%	& 38.5 \%	 & 0.6111                \\
non-filtered     &              &             &                 \\
\hline
self+cat biased   &              &             &                 \\
+ $1$-nbr img tag     &  28.05\% & 39.37\% & 0.6147                \\

+ $1$-nbr  text     &              &             &                 \\
\hline
\end{tabular}
    \caption{VisualTextRank ablation study for ENG-eval.}
    \label{tab:ablation}
\end{table}

For the non-ENG eval set, the ablation study (Table~\ref{tab:ablation_non_eng}) shows that similar ads in English provide useful information for non-English ad text-to-image query task (and hence facilitate cross-lingual learning). In particular, the results follow the same pattern as for the ENG-eval set, except that the best result (line 5 in Table~\ref{tab:ablation_non_eng}) is achieved by augmentation with only similar image tags, and no similar ad text words. 
\begin{table}[h]
    \centering
    \begin{tabular}{|l|c|c|c|c|}
    \hline 
model              & hard         & soft        & avg. w2v \\
                   & accuracy     & accuracy    & similarity     \\
\hline	
\hline
self biased         &              &             &                 \\
TextRank            &     22.39\%  & 29.75\%	 &  0.5235 \\
\hline
cat biased         &              &             &                 \\
TextRank            &     21.78\%  & 28.83\%	 &  0.5409 \\
\hline
self+cat biased      &              &             &                 \\
TextRank        &     24.23\%	& 31.29\%	 & 0.5575          \\
\hline
self+cat biased  &              &             &                 \\
+ $1$-nbr text     & 23.93\% & 31.60\% & 0.554              \\
\hline
self+cat biased   &              &             &                 \\
+ $1$-nbr img tag     &  24.85\%	& 32.21\%	 & 0.5676              \\
\hline
self+cat biased   &              &             &                 \\
+ $1$-nbr img tag     &  24.54\%	& 31.9 \%	 & 0.5608                \\
non-filtered     &              &             &                 \\
\hline
self+cat biased   &              &             &                 \\
+ $1$-nbr img tag     &  23.62\% & 31.60\% & 0.5563                \\
+ $1$-nbr  text     &              &             &                 \\
\hline
\end{tabular}
    \caption{VisualTextRank ablation study for non-ENG-eval.}
    \label{tab:ablation_non_eng}
\end{table}
Table~\ref{tab:examples} shows two (anonymized) examples of the data samples in the non-ENG-eval set with image search queries, and algorithm's output for (self) biased TextRank and VisualTextRank. The first sample in Table~\ref{tab:examples} is an English ad text, and the second sample is a Spanish ad text. In both cases, VisualTextRank returns a keyword matching the (golden) query, unlike the (self) biased TextRank.
\begin{table}[h]
    \centering
\begin{tabular}{|p{2.5cm}|p{2cm}|p{1.3cm}|p{1.3cm}|}
\hline
  ad & user & biased & Visual- \\
   text & queries & TextRank & TextRank \\
 \hline
 \hline
 Online Store. Asian antique and vintage furniture & furniture; wooden furniture & antique & furniture \\
 \hline
 Necesitas Vender Tu Casa Rápido? & home selling; house & sell & house \\
 \hline
\end{tabular}
    \caption{Biased TextRank vs VisualTextRank examples.}
    \label{tab:examples}
\end{table}
For these two ad examples, we also show their (anonymized) nearest neighbours and tags of the corresponding neighbor images (Table ~\ref{tab:examples_nbr}). Note that in both cases neighbour image tags coincide with the ground truth user query and shift algorithm predictions towards the correct output (furniture and house, correspondingly).  
\begin{table}[h]
    \centering
\begin{tabular}{|p{2.5cm}|p{3.5cm}|p{1.3cm}|}
\hline
  ad & nearest & nbr image \\
   text & nbr ad & tag \\
 \hline
 \hline
 Online Store. Asian antique and vintage furniture & Furniture and Decor Sale. Up to 70\% Off Top Brands And Styles! & furniture \\
 \hline
 Necesitas Vender Tu Casa Rápido? & Homeowners Could Sell Their Homes Fast. \_\_\_\_ realtors get the job done. & house; home \\
 \hline
\end{tabular}
    \caption{Examples of ads nearest neighbours and image tags.}
    \label{tab:examples_nbr}
\end{table}
\subsection{Online results}
We conducted an A/B test where advertisers were split into test and control buckets (50-50 split). Test advertisers saw recommended images (via automatic text-to-image search query) when they accessed the image search feature, while control advertisers saw a random set of images.
Both test and control advertisers were permitted to query the stock image library further after seeing the initial set of (recommended or random) images.
Due to latency constraints, a lighter version of VisualTexTRank 
(with category biasing, averaged word2vec embedding for text instead of SBERT, and no augmentation via similar ads) was used. The end-to-end latency of this version was below a second.
In the A/B test, we saw a $28.7\%$ lift in the selection rate in the test bucket (compared to control). In terms of
onboarding rate, we saw a $41.64\%$ lift.
The data was collected over a one month period, and was limited to US-only traffic. Note that, the final image selected from the stock image library need not be from the initial set of images shown to the advertiser. However, the selection rate lift indicates higher adoption of stock images (driven by initial recommendations which increase the advertiser's interest to query further). The lift in onboarding rate is indicative of the positive impact on the ad platform.

%% file: discussion.tex
\section{Discussion} \label{sec:discussion}
The motivation behind VisualTextRank stemmed from insufficient data for supervised training for the ad text-to-image query task. As more advertisers explore stock images (influenced by VisualTextRank), we may collect enough data to fine-tune pre-trained multi-modal models \cite{li2020oscar} on our task.
Cross-lingual learning with such multi-modal models for our task is another future direction.

%% file: reproducibility.tex
\section*{Notes on Reproducibility}
In this section, we list helpful information for reproducing the results in our paper.

\paragraph*{POS tagging and offline metrics} For POS tagging based filters (\emph{i.e.}, to focus on only nouns, proper nouns and verbs), we used Spacy's POS tagger \cite{spacy}. For, the offline evaluation method based on word2vec similarity between the ground truth query and model output we used Spacy's word similarity function implemented using $300$ dimensional word2vec embeddings (trained on Google News dataset) for English text \cite{spacy}.

\paragraph*{TextRank and baselines for offline evaluation}
We used the PKE \cite{pke}
library for running experiments with baseline keyword extraction methods: Position Rank, Topic Rank, Topical Page Rank, Multipartite Rank, and YAKE (as reported in Section~\ref{sec:ENG-eval}).
For TextRank and biased TextRank we used the implementation in
\url{https://github.com/ashkankzme/biased_textrank}, with damping factor $d=0.8$, iterations = $80$ (for the node importance estimate to converge), and node similarity threshold = $0.9$ (using SBERT embeddings).
The SBERT \cite{sbert_paper} embeddings were based on stsb-distilbert-base as listed in the pre-trained models in \url{https://www.sbert.net/docs/pretrained_models.html}.
We explored the larger BERT models like BERT-base, BERT-large, RoBERTa-large but the benefits were not significant.

\paragraph*{Category list for computing category bias}
We used a list of $\sim 250$ categories as the set $\mathbf{C}$ defined in Section~\ref{sec:category_biasing}. Each entry in $\mathbf{C}$ is a phrase corresponding to a category in the flattened IAB ads category taxonomy \cite{iab_mopub}. For example, the `Arts \& Entertainment' high level IAB category has `Books \& Literature' and `Music' as subcategories \cite{iab_mopub}. After flattening, `Arts \& Entertainment', `Books \& Literature' and `Music'
are listed as three separate entries in category set $\mathbf{C}$ used for computing category bias.

\paragraph*{VisualTextRank}
For the best results from VisualTextRank we used the following configuration (for both ENG-eval and non-ENG-eval):
\begin{itemize}
    \item damping factor $d = 0.8$
    \item node similarity threshold = $0.9$ (using distilbert based SBERT embeddings)
    \item min. tag similarity for augmentation $min\_tag\_sim = 0.7$ 
    \item min. word similarity for augmentation $min\_word\_sim = 0$ \item for category biasing, we set all $bias_i^{(cat)}$ below $0.4$ to zero.
\end{itemize}
Apart from using distilbert based SBERT, we also explored the larger BERT models like BERT-base, BERT-large, RoBERTa-large but the benefits were not significant.
For the multilingual ad experiment with xlm-SBERT
(\emph{i.e.}, without ad text translation) described in Section~\ref{sec:multilingual}, we used the stsb-xlm-r-multilingual model which was trained on parallel data for 50+ languages as mentioned in \url{https://www.sbert.net/docs/pretrained_models.html}.